\documentclass[letterpaper, 10pt, conference]{ieeeconf}  

\IEEEoverridecommandlockouts                              
                                                          
\usepackage[utf8]{inputenc}
\usepackage[english]{babel}
\usepackage{amsmath}
\usepackage{amssymb}
\usepackage{graphicx}
\usepackage{subcaption}
\usepackage{hyperref}
\usepackage{url}
\usepackage{xcolor}

\usepackage{geometry}
\IEEEaftertitletext{\vspace{-1.0\baselineskip}}

\usepackage[nocompress]{cite}

\usepackage{dcolumn,booktabs}
\newcolumntype{d}[1]{D{.}{.}{#1}} 

\newcommand{\doble}[2]{\begin{tabular}{@{}c@{}}#1\\#2\end{tabular}}
\newcommand{\triple}[3]{\begin{tabular}{@{}c@{}}#1\\#2\\#3\end{tabular}}

\DeclareMathOperator*{\argmin}{arg\ min} 

\newcommand{\norm}[1]{\left\lVert#1\right\rVert}

\title{\LARGE \bf Scale-aware direct monocular odometry}
\author{Carlos Campos and Juan D. Tardós,~\IEEEmembership{Fellow,~IEEE}
\thanks{This work was supported in part by the Spanish government  grant  PGC2018-096367-B-I00 and 
the Aragón government  grant DGA\_T45-17R.
}
\thanks{The authors are with Instituto de Investigación en Ingeniería de Aragón (I3A), Universidad de Zaragoza, Spain   {\tt\small campos@unizar.es; tardos@unizar.es}}%
}

\begin{document}

\thispagestyle{empty}
\newpage
\onecolumn
\begin{center}
This paper has been accepted for publication in 2022 International Conference on Intelligent Robots and Systems (IROS).
\vspace{0.75cm}\\
DOI: \\ 
IEEE Xplore: \\
\vspace{1.25cm}
\end{center}
©2022 IEEE. Personal use of this material is permitted. Permission from IEEE must be obtained for all other uses, in any current or future media, including reprinting/republishing this material for advertising or promotional purposes, creating new collective works, for resale or redistribution to servers or lists, or reuse of any copyrighted component of this work in other works.
\twocolumn

\maketitle
\thispagestyle{empty}
\pagestyle{empty}

\begin{abstract}
We present a generic framework for scale-aware direct monocular odometry based on depth prediction from a deep neural network. In contrast with previous methods where depth information is only partially exploited, we formulate a novel depth prediction residual which allows us to incorporate multi-view depth information. In addition, we propose to use a truncated robust cost function which prevents considering inconsistent depth estimations. The photometric and depth-prediction measurements are integrated into a tightly-coupled optimization leading to a scale-aware monocular system which does not accumulate scale drift. Our proposal does not particularize for a concrete neural network, being able to work along with the vast majority of the existing depth prediction solutions. We demonstrate the validity and generality of our proposal evaluating it on the KITTI odometry dataset, using two publicly available neural networks and comparing it with similar approaches and the state-of-the-art for monocular and stereo SLAM. Experiments show that our proposal largely outperforms classic monocular SLAM, being 5 to 9 times more precise, beating similar approaches and having an accuracy which is closer to that of stereo systems. 

\end{abstract}

\section{Introduction}
In the last years, advances in  machine learning have shaken the entire computer vision community. Compared with classical methods, learning based solutions have proved to be outstanding in some tasks like object detection, scene representation or monocular depth prediction. These have direct application to the Simultaneous Localization and Mapping (SLAM) problem which have been and continue to be studied by the community.

With respect to single-view depth prediction, proposed Convolutional Neural Networks (CNN) are able to accurately estimate pixelwise depth for images close to the training domain. Having such an estimation allows pure monocular SLAM/odometries \cite{mur2015orb,Engel-et-al-pami2018} to estimate the true scale of the map, as stereo \cite{mur2017orb} or visual-inertial \cite{campos2021orb} systems do. In addition, it mitigates and removes most important pure monocular issues like scale drift \cite{strasdat2010scale} or need of an ad-hoc map initialization process. Furthermore, for autonomous driving situations, using a monocular-inertial odometry may not be feasible. Vehicles do not perform 6DoF motion, making inertial parameters, such as IMU biases or scale, have low or null observability.

In this work we leverage these latest advancements and propose a direct monocular odometry pipeline, to tightly integrate information from intensity images and depth prediction inferred from existing CNNs, building {\color{black}general} a scale-aware system. Using as inputs only the intensity image and the predicted depth allows us to use a large set of existing neural networks, increasing the applicability of our proposal.

\begin{figure}
    \begin{subfigure}{\columnwidth}
        \centering
        \includegraphics[width=0.95\textwidth]{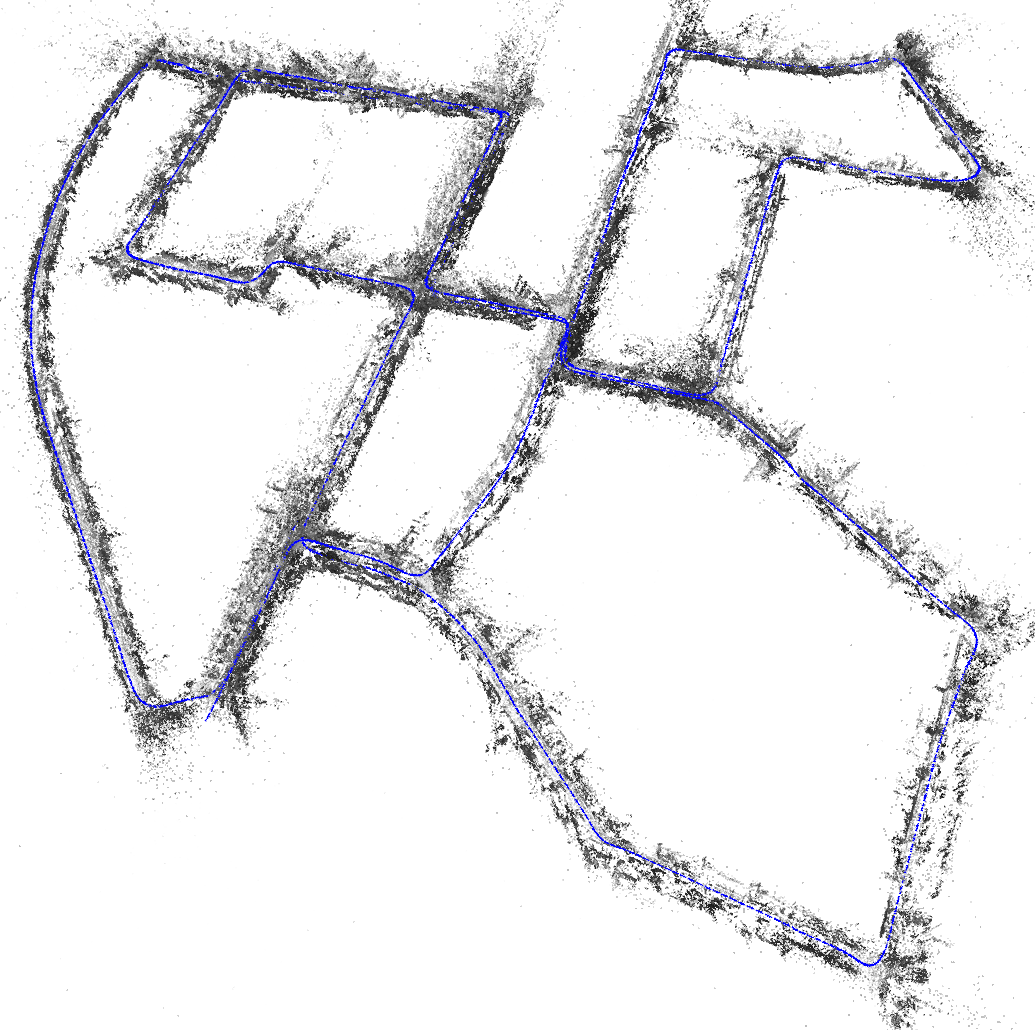}
        \label{subfig:KITTI0_pointcloud}
    \end{subfigure}

    \begin{subfigure}{\columnwidth}
      \centering
      \includegraphics[width=0.85\textwidth]{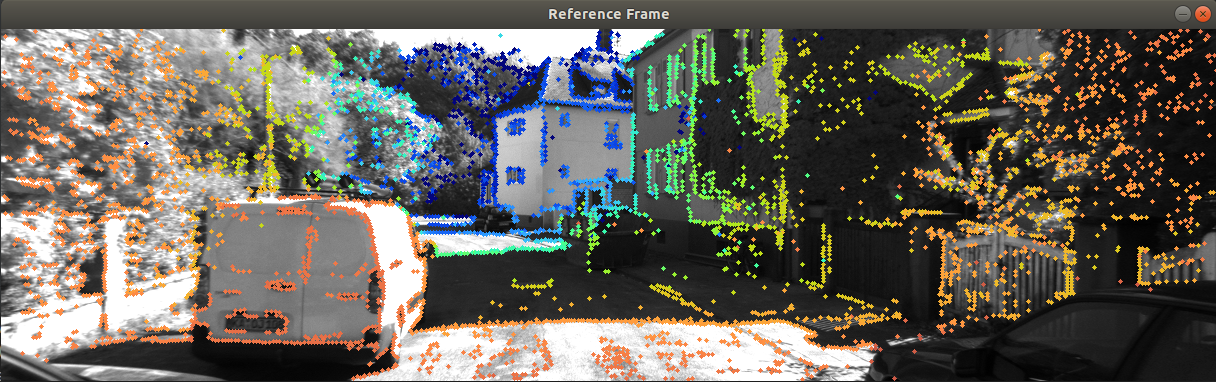}
      \label{subfig:KITTI0_depthmap}
    \end{subfigure}
    \caption{Our reconstructed point-cloud and estimated sparse depth map from visual odometry for KITTI00. {A video for this sequence is available at \url{https://www.youtube.com/watch?v=mZS-gaVYxEU}}  }
    \label{fig:teaser_image}
\end{figure}

\vfill\null
Next, we enumerate the main contributions of our work,
\begin{itemize}
    \item A novel tightly-coupled optimization for photometric and depth prediction measurements. In contrast with previous work \cite{tateno2017cnn, yang2018deep, yang2020d3vo}, depth prediction residuals are formulated independently of the intensity image and are included for all observer frames, not only the first one ({\em host} / {\em anchor}). This allows us to use image points with lower intensity gradient and fully exploit depth prediction measurements, without an increment in computational cost.
    \item A robust optimization which makes use of \textit{Truncated Least Square} (TLS) cost \cite{yang2020graduated} for depth-prediction residual. This prevents considering inconsistent depth measurements during optimization.
    \item A general system that can be used with any existing or future depth prediction neural network, {\color{black} as shown in the results}, performing better than the similar solution DF-VO \cite{zhan2020visual}.
\end{itemize}

\section{Related Work}

The scene depth remains ambiguous from monocular images and can not be directly recovered from a single camera. However, it is clear that there exists some relation between intensity image and image depth. Given a large enough dataset, learning based methods are able to learn this relation, allowing us to infer pixel's depth from gray scale images. 

 One of the first successful works on single-view depth estimation was presented by Eigen et al. \cite{eigen2014depth}. They proposed a CNN with two components, one for the global structure of the image, and the other one to recover fine details, being trained in a semi-supervised way. Latter work from Godard et al. \cite{godard2017unsupervised} extended the used of CNNs for this task with some important improvements. They formulated different loss functions which allowed unsupervised training, extending the applicability of this approach. More recent work CAM-Convs \cite{facil2019cam} from Fácil et al. generalizes the solution to different camera {intrinsics}. They present a new kind of convolution which sidesteps the need of training from scratch a new network when camera parameters are modified. In this work we will use an evolution of \cite{godard2017unsupervised}, coined as \textit{monodepth2} \cite{godard2019digging}. This presents some improvement regarding occlusion and outliers detection, and its implementation is publicly available. {\color{black} Another very recent opensource method, {\em GLPDepth} \cite{kim2022global}, leverages on transformer encoders, aiming to improve the long-range dependencies of depth values along the image predictions}

The first time depth prediction from a CNN was used for SLAM or odometry was at CNN-SLAM \cite{tateno2017cnn} by Tateno et al, where consecutive frames were aligned using photometric and depth prediction measurements, {borrowing ideas from RGB-D systems}. Once obtained these relative transformations, the entire set of keyframe poses was optimized in a pose-graph fashion. More recent solutions include DVSO \cite{yang2018deep} and D3VO \cite{yang2020d3vo} by Yang et al. Both are carefully designed to make them work specifically with DSO \cite{Engel-et-al-pami2018}, which is used as the odometry system as well as for the neural-network supervised training. At DVSO a neural network is trained to estimate not only pixel's depth, but also the disparity in the virtual right camera, which is also used as an odometry input to discard inconsistent points. At D3VO, depth prediction uncertainty, relative pose and brightness transformations between consecutive frames are also computed. This allows to accordingly weight measurements and include pose-graph constraint in the photometric Bundle Adjustment. This approach obtains outstanding results which are similar, if not better, than state-of-the-art stereo solutions. At DF-VO \cite{zhan2020visual} a more general approach is followed. Depth prediction neural network and its respective odometry are decoupled, being both independently developed. This allows to combine that odometry with any existing or future depth prediction module, gaining flexibility.

In this work we follow a similar approach to DF-VO, in contrast with DVSO or D3VO where the depth prediction module and the odometry system are tightly related. Differently from DF-VO, which also estimates optical flow between consecutive images, our proposal only requires predicted depth as input, keeping our solution even more general {\color{black} and making it ready to work by just plugging any depth prediction neural network. In figure \ref{fig:system_overview} we show the overview of our proposal, which is described in detail in the next section.}

\section{Direct visual odometry with depth prediction}
\begin{figure}
  \centering
  \includegraphics[width=1.0\columnwidth]{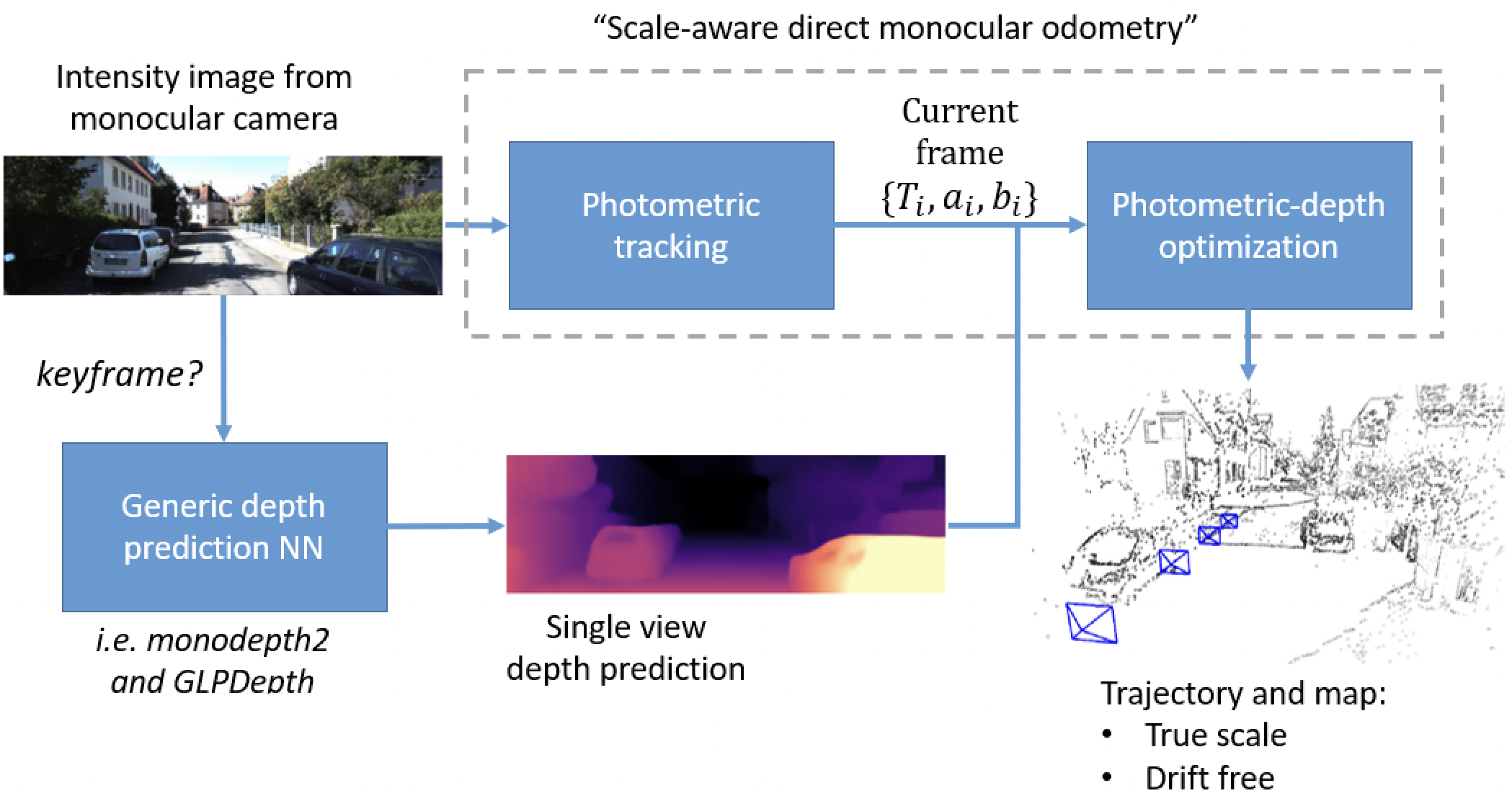}
  \caption{System overview.}
  \label{fig:system_overview}
\end{figure}

In this section we describe our proposal, presented in  figure \ref{fig:system_overview}. The intensity image is used for photometric tracking to compute the current frame state. If a new keyframe is inserted, depth is estimated from a generic depth prediction neural network. Photometric and depth prediction information are combined together to run a photometric-depth optimization leading to a trajectory and map drift free and with the true scale. Block elements inside the dashed box corresponds to our proposal, which is completely agnostic of any other component out of the box.

The map structure for our odometry consists of map points and keyframes. Map points are represented using an inverse depth parametrization \cite{civera2008inverse}. For the first observer keyframe, named as {\em host} or {\em anchor}, we keep azimuth and elevation fixed, having only inverse depth $\rho$ as optimizable parameter \cite{Engel-et-al-pami2018}. A keyframe will be parametrized with its pose $\mathbf{T} \in \text{SE}(3)$ and its brightness affine transformation $(a,b)$. For tracking purposes, we define a temporal active window consisting of the last $N_a=5$ keyframes. All points seen from these keyframes will define the set of active map points, which are those probably observed from current frame. We also define a larger optimization window, consisting of the $N_o=7$ last keyframes. These will be optimized in a back-end photometric-depth Bundle Adjustment along with map points observed or hosted by them.

In the following points we will describe in detail our proposal. First, at \ref{subsec:data_management}, we explain how map points and keyframes are created and removed. We continue with the front-end photometric tracking, at point \ref{sec:photometric_tracking}. Finally, at \ref{sec:Photometric-Depth-optimization}, we present our novel back-end photometric-depth optimization.

\subsection{Map point and Keyframe management}
\label{subsec:data_management}

In contrast with pure monocular approaches, having a single-view depth estimation allows to initialize points just from one view. This is also especially important for map initialization, since we can boost our system with a single frame, as stereo odometries do. The keyframe where the point is initialized will be set as the {\em anchor} or {\em host} keyframe. When a new keyframe is inserted, we initialize points in image regions where there exist no observations. To this end, we compute the mean $\mu$ and standard deviation $\sigma$ of intensity values at each cell along a $16 \times 32$ grid in the image. For each cell, we extract points whose gradient is above $\mu + f \sigma$, where $f$ is an adaptive factor. We make several extractions, starting with a high $f = 10$ value and decreasing it until we have at least 2000 hosted or observed points in the current keyframe. Each time we extract a new point, we mask its neighbors within a $5 \times 5$, window to avoid extracting overlapping points.


We also perform a map point culling process. In this way, a map point can be removed from the map for three reasons:
\begin{enumerate}
    \item When a map point has been created outside of the active window and it has less than 2 observations. This removes points which are difficult to track.
    \item When the mean photometric residual of its observations is higher than 9 intensity values. This removes probably bad estimated points.
    \item When it has at least one observation and its inverse depth information, quantified with Hessian block from previous BA, drops below a threshold. In this sense even a point with several observations may contain little information. This is the case when its geometric derivatives (very far points) or image gradient (textureless points) are small, or they are orthogonal to each other (see appendix).
\end{enumerate}

For keyframe creation we follow a simple heuristic: when the number of inliers in frame tracking (see section \ref{sec:photometric_tracking}) drops below 70\%, a new keyframe is created. 
Each time a keyframe is inserted, all active points are assumed to be observed. It is the photometric-depth optimization which will manage and discard outlier observations, as explained in \ref{sec:Photometric-Depth-optimization}. When a keyframe gets out of the active window and more than 80\% observed and hosted points have at least 3 observations, we discard that keyframe since we consider it contains too redundant information.

\subsection{Photometric Tracking}
\label{sec:photometric_tracking}
This task consists in estimating the current frame state, relative pose and brightness affine transformation with respect to the last keyframe, named as reference keyframe. We will follow a procedure similar to DSO \cite{Engel-et-al-pami2018}.

For this goal, when the reference keyframe, with image $\Omega$, is updated, we first build a sparse depth map, $D: \Omega_D \xrightarrow{} \mathbb{R}$, by projecting active points into $\Omega$, and dilating them to get a denser set $\Omega_D \subset \Omega$. Since map points are created from a single view, projected points may have from one (low accuracy) to multiple (high accuracy) observations. In contrast with DSO and to take into account this disparate amount of information and accordingly weight each projected point, we propose to weight them using the information value of its inverse depth, computed from the last photometric-depth bundle adjustment. This limits the influence of recently created points with higher uncertainty while points with more observations and better conditioned will dominate the solution. 

We remark we build this depth map $D$ from our estimated point cloud instead of directly using the predicted depth map from the neural network. Since our estimated points have been refined combining photometric and depth-prediction residuals, they are much more precise than network output. Once map $D$ is built, we solve the following optimization problem:
\begin{multline}
\label{eq:photometric_tracking}
\argmin_{\{\mathbf{T}, a, b\}_{i,\text{ref}}} \sum_{\mathbf{u} \in \Omega_{D}} \rho_{\text{Hub}}\left(\left\lVert I_{\text{ref}}(\mathbf{u}) -  \right. \right. \\  \left. \left. e^{-a_{i,\text{ref}}}\{I_i(\pi(\mathbf{T}_{i,\text{ref}}\pi^{-1}(\mathbf{u},D(\mathbf{u}))))-b_{i,\text{ref}}\}\right\lVert^2\right)
\end{multline}

Where grey-scale images are defined as ${I}:\Omega \xrightarrow[]{} [0,255]$ and $\pi : \mathbb{R}^3 \xrightarrow[]{} \Omega$ and $\pi^{-1} : \Omega \times \mathbb{R} \xrightarrow[]{} \mathbb{R}^3$ are the camera projection map and its inverse. Unknowns to be found are $\mathbf{T}_{i,\text{ref}} \in \text{SE}(3)$, the transformation from reference keyframe to current frame, and $\{a_{i,\text{ref}}, b_{i,\text{ref}}\}$, its relative brightness affine transformation, as explained in \cite{Engel-et-al-pami2018}. Relative motion is initialized assuming a constant velocity model, while affine parameters are set to values from the last frame. In addition, we use a Huber robust norm {\cite{mactavish2015all}} to downweight outlier observations. We run this optimization in a multiscale fashion { \cite{bouguet2001pyramidal}}, starting at the coarsest scale level and going down until the original image resolution.{In the implementation we use five levels with a scale factor 2.} We optimize with Levenberg-Marquardt until convergence at each level, with a maximum of 20 iterations per level, since this optimization is very efficient. If we detect that the optimization has not converged in the coarsest level, usually due to big rotations, we update the initial attitude estimate with {$\pm 10^{\circ}$ rotations around each of the axes and run the optimization again. Among the found solutions, }we keep the one which provides the lowest mean photometric residual. Along this optimization we do not include depth residuals which avoids running the depth prediction at frame rate, bringing an important computational saving.

Instead of solving (\ref{eq:photometric_tracking}), we follow the more efficient inverse compositional approach as presented in \cite{baker2004lucas}, also adopted in DSO.


\subsection{Photometric-Depth optimization}
\label{sec:Photometric-Depth-optimization}
\begin{figure}
  \centering
  \includegraphics[width=1.0\columnwidth]{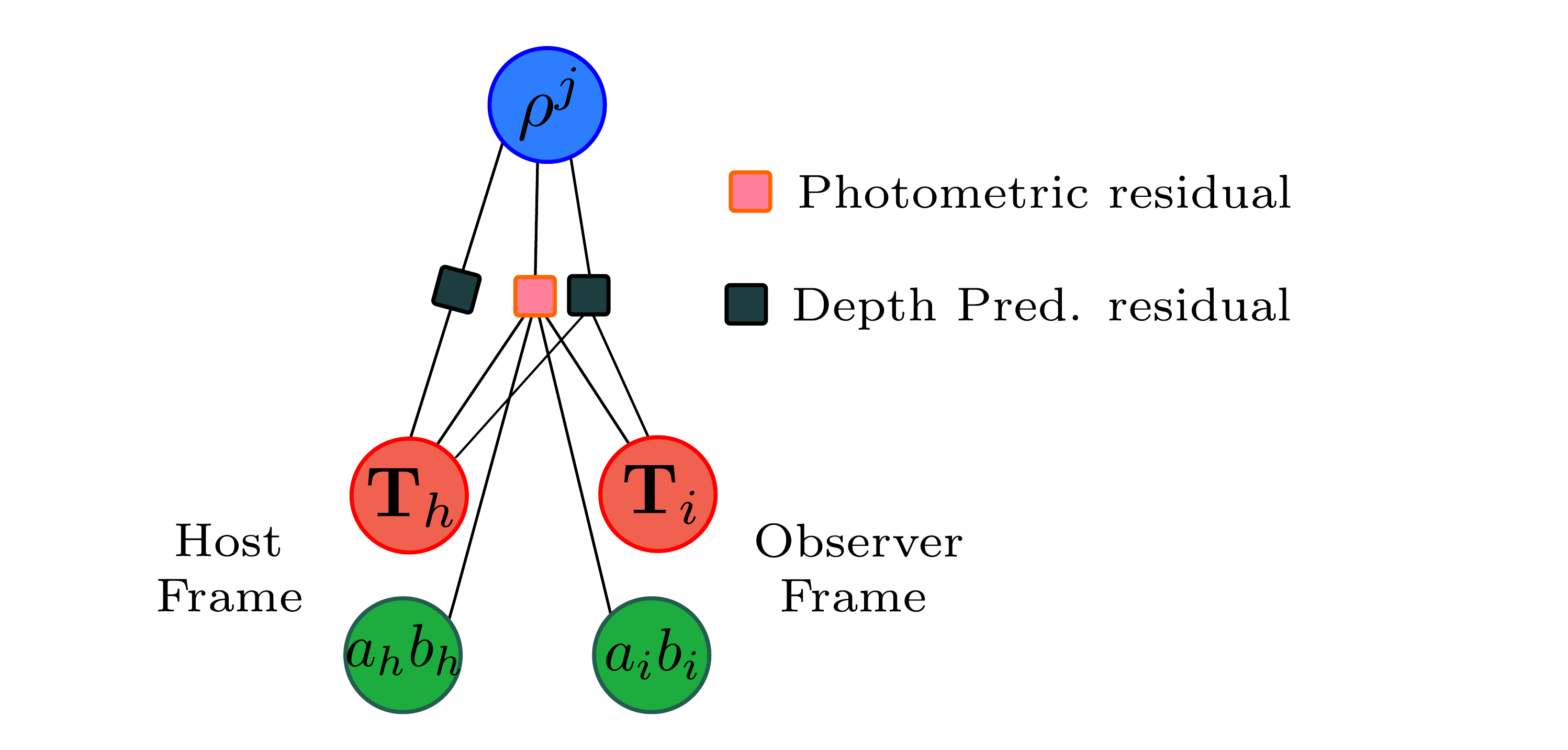}
  \caption{Simplified factor graph representation for photometric-depth BA with one point, its host and one observer.}
  \label{fig:photo_depth_BA}
\end{figure}

Each time a new keyframe is inserted, we run a windowed optimization for the last $N_o=7$ keyframes, denoted as $\mathcal{K}$. Variables to be optimized are keyframe poses and affine parameters inside the optimizable window, as well as inverse depth from all their observed points $\mathcal{P} = \{\mathcal{P}_1 \cup \dots \cup \mathcal{P}_{N_o}\}$. All observations for these points are included while observer and host keyframes outside the optimization window remain fixed. For this optimization we include two kinds of residuals, with a parameter $k$ weighting between them ($5\times 10^3$ in our implementation). This leads to the following optimization problem:
\begin{equation}
    \argmin_{\{\mathbf{T}_i,\rho^j\}_{i, j}} \sum_{i \in \mathcal{K}} \sum_{j \in \mathcal{P}_i} \rho_{\text{Hub}}\left(\norm{{r}_{i,\text{photo}}^j}^2\right) +  k^2 \, \rho_{\text{TLS}}\left(\norm{{r}_{i,\text{depth}}^j}^2\right)
\end{equation}
A simplified factor graph representation for this optimization problem is given in figure \ref{fig:photo_depth_BA}.

The first residual, ${r}_{i,\text{photo}}^j$, is the photometric one, which relates point $j$ and observer keyframe $i$ as follows,
\begin{equation}
\label{eq:photometric_pattern_residual}
    {r}^j_{i,\text{photo}} = \sum_{{\mathbf{u}^j_h}' \in \mathcal{N}_{\mathbf{u}_j}} I_h\left({\mathbf{u}^j_h}'\right) - b_h - \frac{e^{a_h}}{e^{a_i}} \left(I_i\left({\mathbf{u}^j_i}'\right) - b_{i}\right)
\end{equation}
such that
\begin{equation}
\label{eq:image_point_transformation}
    {\mathbf{u}^j_i}' = \pi\left({\mathbf{x}^j_i}'\right) \quad \text{with} \quad {\mathbf{x}^j_i}' = \mathbf{T}_{ih}  \pi^{-1}\left({\mathbf{u}^j_h}',\rho^j\right)
\end{equation}
Where $\rho^j$ is the inverse depth and ${\mathbf{u}^j_h}'$ are the image coordinates of neighbor pixels in the host for point $j$. We use the same patch $\mathcal{N}_{\mathbf{u}_j}$ as proposed at DSO \cite{Engel-et-al-pami2018} which allows fast vectorized computation. In contrast with DSO, where the $\mathbf{T}_i$ update is performed in the local reference, we prefer to apply it on the global reference. This makes derivatives with respect to the host and observer frame only differ on sign, as shown in the appendix, resulting in a computational reduction. The drawback of this formulation is the derivatives depend on distance to origin, which may cause stability issues when far from it. To remove this, we apply a translation offset to the entire set of keyframes, bringing the last keyframe to the origin. Since keyframes involved in optimization are spatially close, this issue disappears. Once solved the optimization, we take all keyframe poses back to the original reference. We use a Huber kernel to linearly weight outliers, with its threshold set to 9 for pixel.

The second residual accounts for depth prediction measurements. From the neural network, we have an inverse depth estimation $D^i_{NN}:\Omega \xrightarrow{} \mathbb{R}^{+}$ for a given keyframe $i$. For each point $j$ observed from keyframe $i$ we define the following depth prediction error:
\begin{equation}
\label{eq:depth_pred_res}
    {r}_{i,\text{depth}}^j = D^i_{NN}(\mathbf{u}^j_i) - \rho^j_i \quad \text{s.t.} \quad \rho^j_i = [\mathbf{x}_i^j]_z^{-1}
\end{equation}
where $\mathbf{x}_i^j$ and $\mathbf{u}^j_i$ can be computed similar to (\ref{eq:image_point_transformation}). $[\cdot]_z$ takes the $z$ component of the vector. For this error we do not use a patch of pixels, since depth map is usually much more smooth than intensity image and close pixels usually contain redundant depth information.

In contrast with RGB-D SLAM systems, that use reliable depth measurements, we use a \textit{Truncated Least Square} (TLS) cost function \cite{yang2020graduated}, also known as threshold cost \cite{mactavish2015all}, to handle depth outliers. When the depth prediction residual is above a threshold, its gradient vanishes, which may be seen as setting its weights to zero, removing its influence. There are two reasons for using this TLS cost. First, when photometric and depth prediction measurements do not agree, we rely on the former and neglect the later which is more prone to inconsistencies. Second, for each map point we have prediction-depth residuals from each observer keyframe, which may not be consistent between them. Prediction accuracy may depend on the point of view, on the region of the image where the prediction is made or on how far the point is. Using a TLS robust cost function allows to activate only depth measurements which are consistent between them or agree with photometric information. The threshold of this TLS cost function is set to 0.01 $m^{-1}$.
We also define the depth prediction residual for the host keyframe, which takes a simpler form,
\begin{equation}
    r^j_{h,\text{depth}} = D_{NN}^h(\textbf{u}_h^j)-\rho^j
\end{equation}

The depth residual proves to be very useful to constraint points with lower photometric information, {namely those which do not have enough intensity gradient or are orthogonal to its epipolar line, avoiding their singular Hessians}.
We remark we are including depth prediction residuals for all observer keyframes, which contrasts with \cite{yang2018deep, yang2020d3vo}, where only host depth measurement is considered. Including this error does not suppose a big computational cost increment, since a lot of terms may be reused from photometric residual (see appendix).

Affine parameters $a$ and $b$ are prone to drift during optimization and they also add extra degrees of freedom to the optimization problem. To avoid this, in addition to these residuals, we add a strong prior to keep them close to zero. As for photometric tracking, this optimization is solved in a multiscale way. However, since previous optimizations have been run for keyframes and map points, estimates are close to the minima. Thus, it is not necessary to start the optimization from the coarsest level. Instead, we start at third finest level, which gives a basin of convergence of 4 pixels.

Once solved the optimization, we discard outlier observations based on two criteria. First, if the average photometric residual for pixels in $\mathcal{N}_{\mathbf{u}_j}$ is over a threshold, 9 intensity values in our implementation, we discard it. Second, if the number of pixels in $\mathcal{N}_{\mathbf{u}_j}$ with a high residual (15 intensity values or more in our implementation) is greater than 40\%, we discard that observation.

{If we compare how photometric and depth prediction residuals are combined together, we find differences with previous works}. At CNN-SLAM \cite{tateno2017cnn}, depth prediction residual is not explicitly used. Instead, photometric residual is made dependent on the depth map prediction. DF-VO \cite{zhan2020visual} solves a PnP problem, using predicted depth. Other methods like DVSO \cite{yang2018deep} or D3VO \cite{yang2020d3vo} convert the inverse depth prediction to an equivalent stereo observation, defining a photometric error for the virtual stereo system. For all these approaches, the residual and its derivatives depend on intensity image and its gradient, which is more prone to contain noise and is much less smooth than predicted depth gradients. In addition, it restricts the use of predicted depth to regions with high visual texture. 

\begin{figure}
    \centering
        \begin{subfigure}{0.44\columnwidth}
      \centering
      \includegraphics[width=1.1\linewidth]{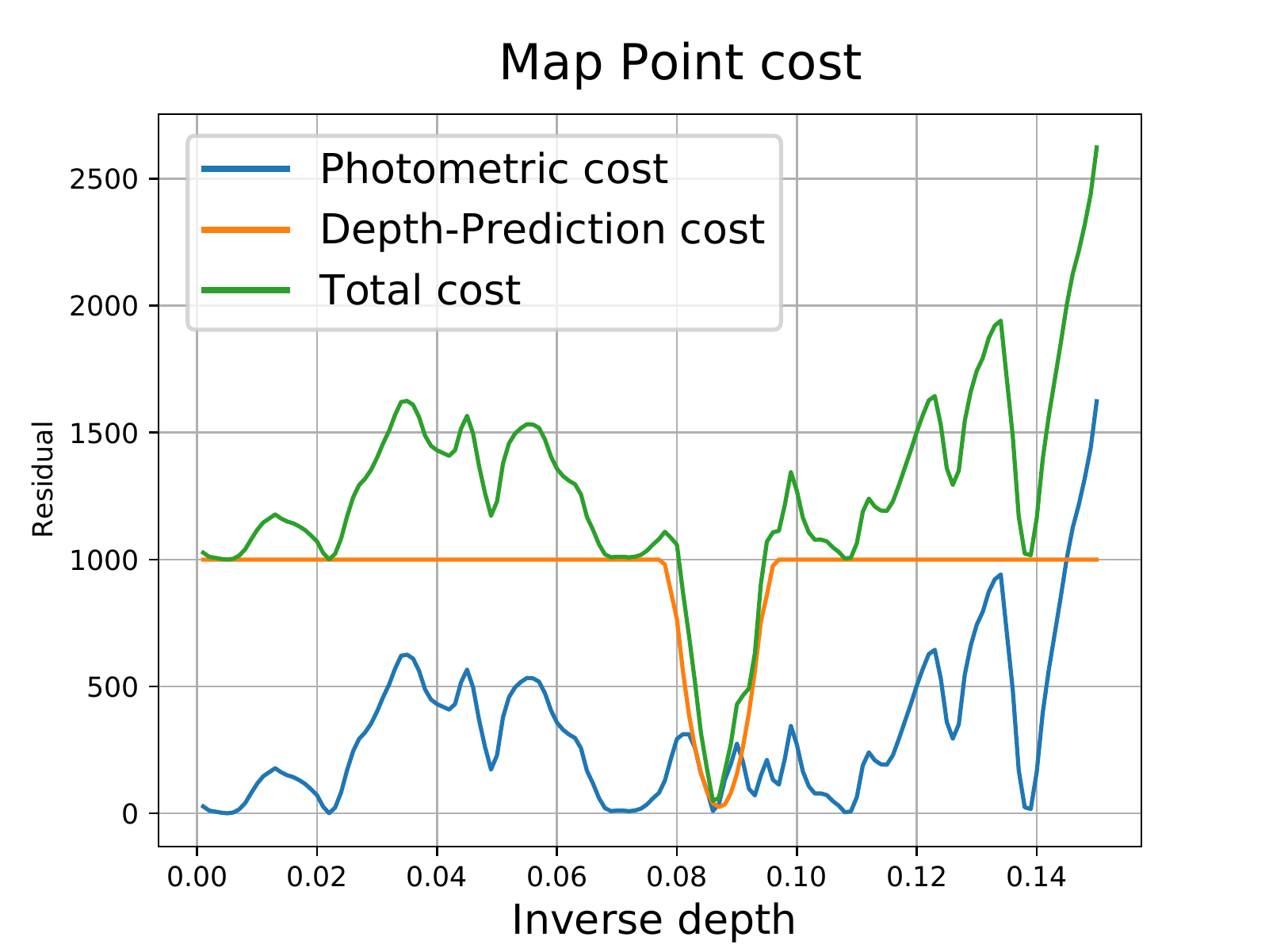}
      \caption{}
      \label{subfig:photo_pred_cost_lowgradient}
    \end{subfigure}
    \begin{subfigure}{0.44\columnwidth}
      \centering
      \includegraphics[width=1.1\linewidth]{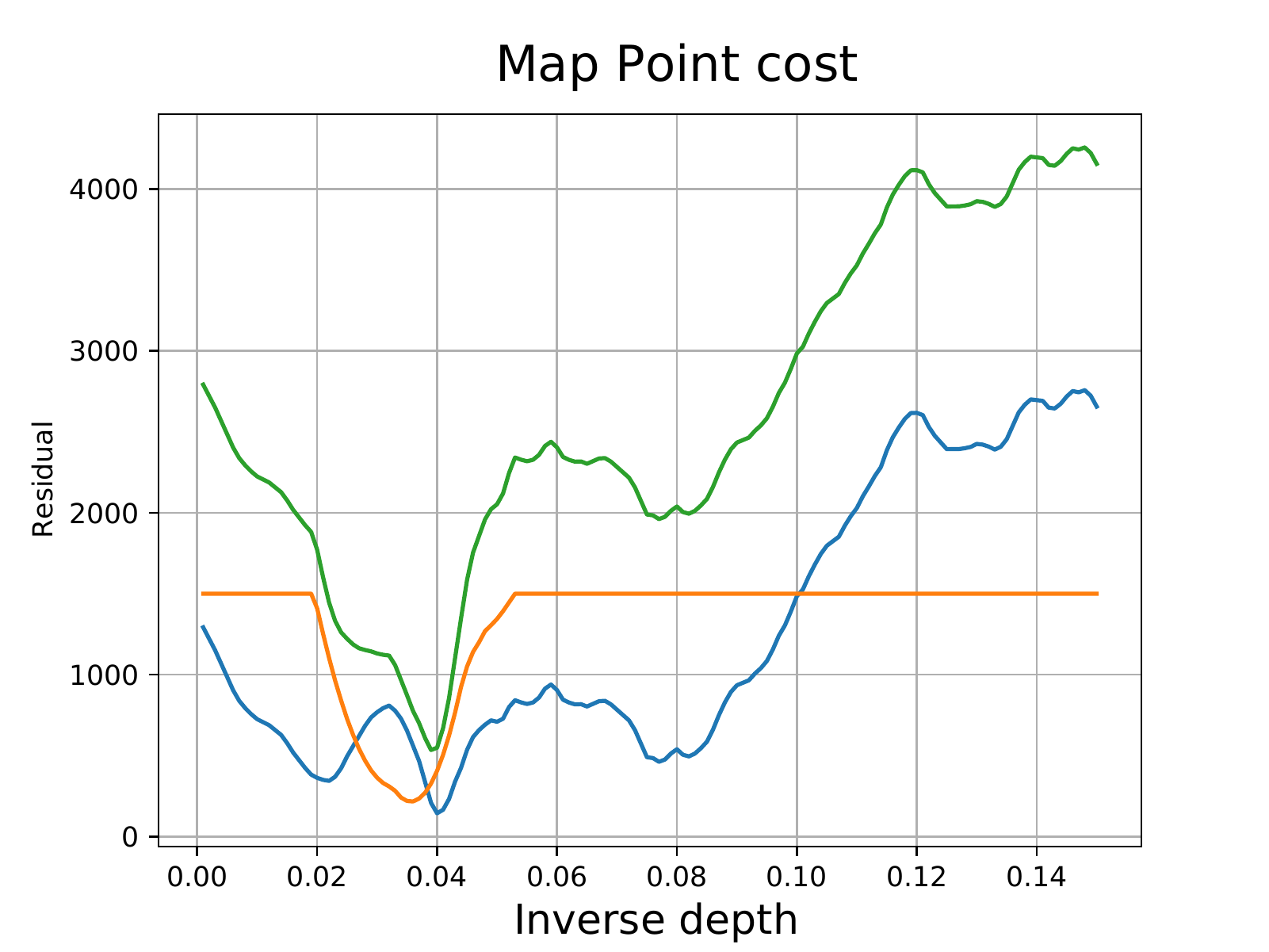}
      \caption{}
      \label{subfig:photo_pred_cost_convexity}
    \end{subfigure}
    
    \begin{subfigure}{0.44\columnwidth}
      \centering
      \includegraphics[width=1.1\linewidth]{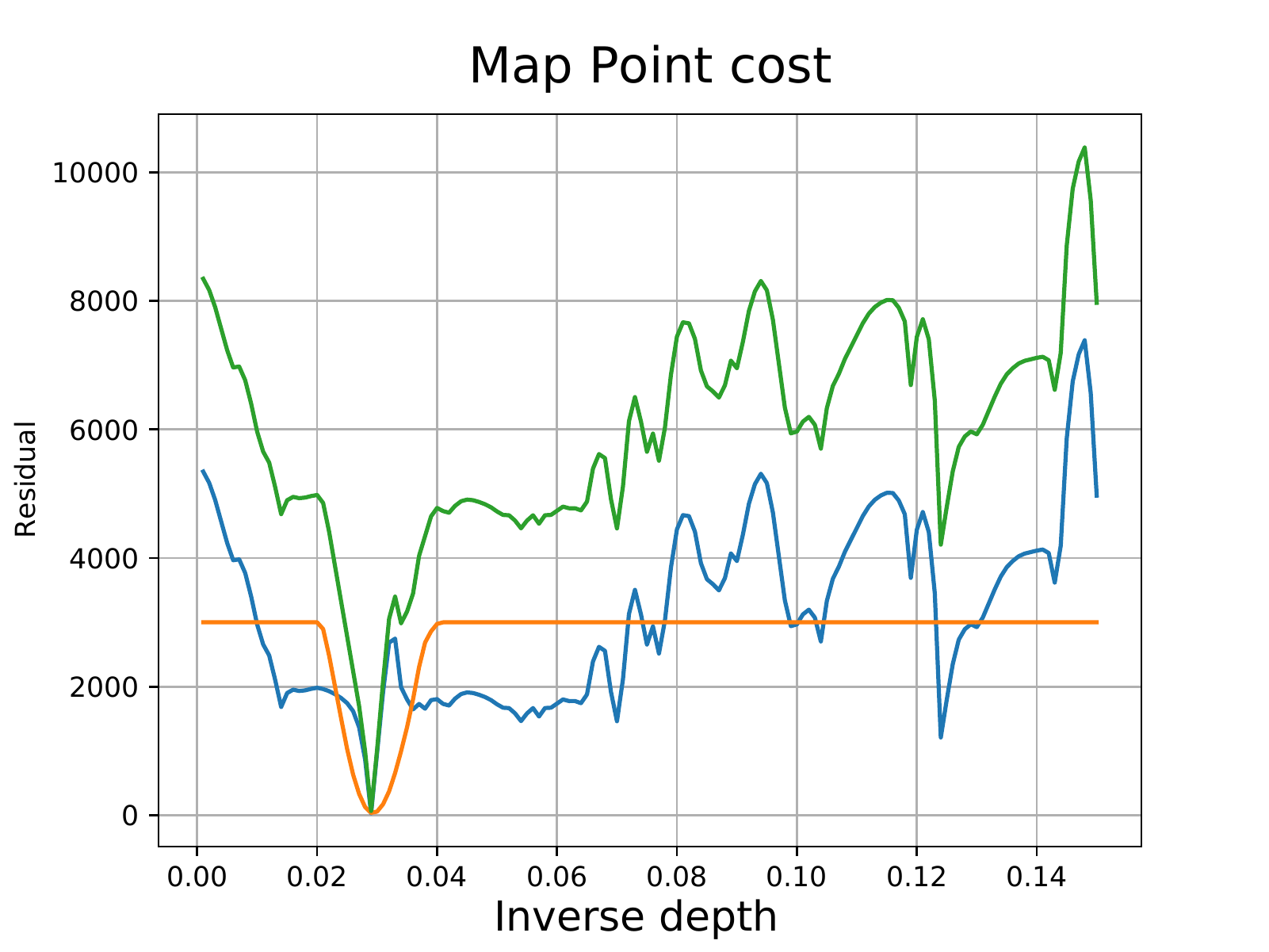}
      \caption{}
      \label{subfig:photo_pred_cost_consistent}
    \end{subfigure}
    \begin{subfigure}{.44\columnwidth}
      \centering
      \includegraphics[width=1.1\linewidth]{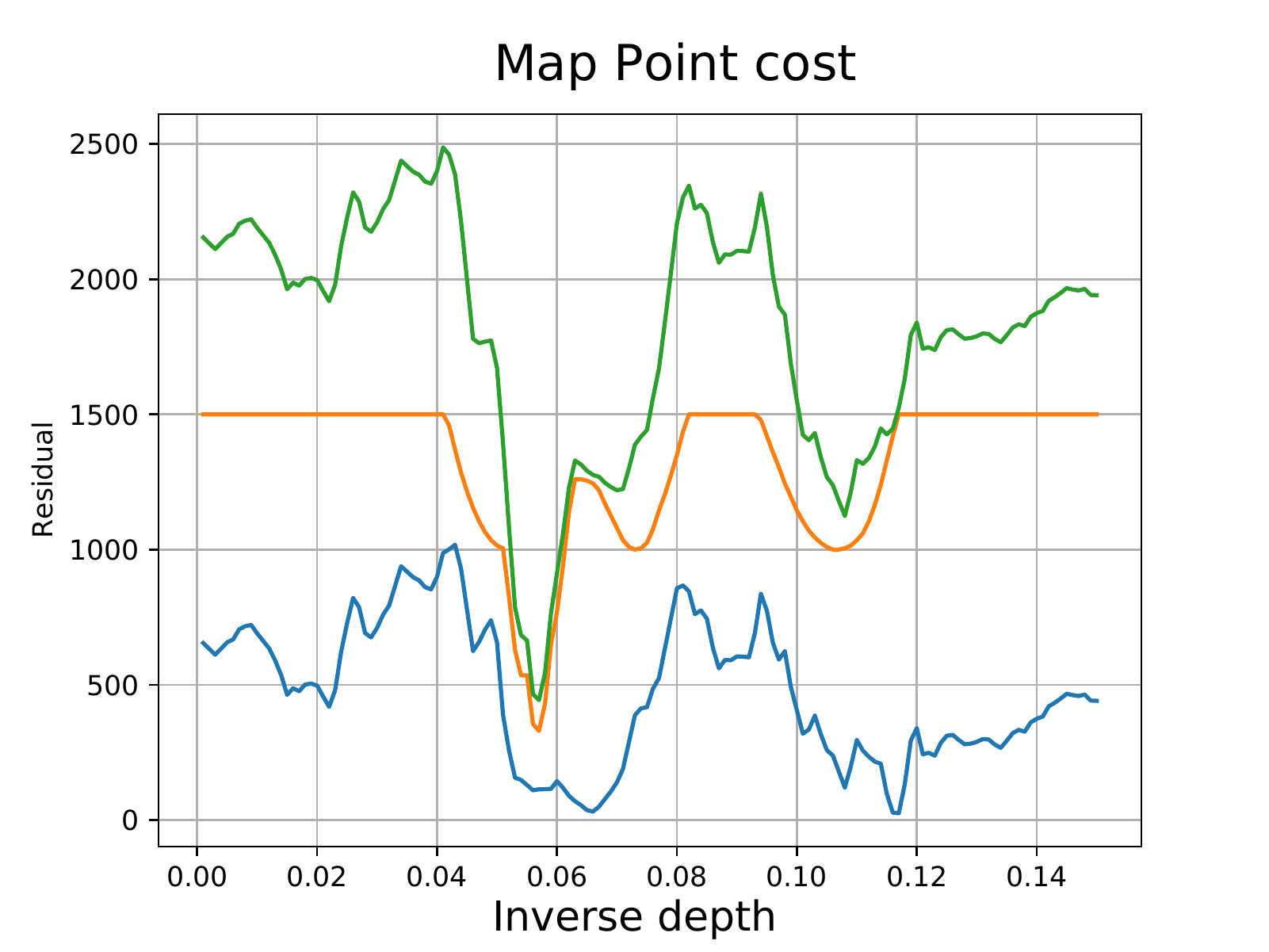}
      \caption{}
      \label{subfig:photo_pred_cost_multiple}
    \end{subfigure}
 
    \caption{Different map points cost w.r.t. its inverse depth.}
    
    \label{fig:cost_evolution}
\end{figure}

In figure \ref{fig:cost_evolution}, we plot photometric, depth prediction and total costs for some randomly chosen map points during photometric-depth BA. In figure \ref{subfig:photo_pred_cost_multiple}, predicted depth cost (orange) has multiple minima since not all predicted depths are consistent. However, the lowest one also corresponds with one photometric minimum (blue), leading to a clear absolute minimum in the total cost (green). Adding predicted depth also increases convergence region of the total cost (Fig. \ref{subfig:photo_pred_cost_convexity}). In some cases, all depth predictions and photometric measurements are consistent with an equivalent minimum, as shown in figure \ref{subfig:photo_pred_cost_consistent}. As we have previously stated, for low gradient points or repetitive regions, whose photometric cost may be plagued with multiples shallow minima, predicted depth increases convergence region for point's depth (Fig. \ref{subfig:photo_pred_cost_lowgradient}).

\section{Results}
We evaluate our proposal on KITTI odometry \cite{geiger2013vision}, a self-driving oriented dataset, using two available neural netoworks: \textit{monodepth2} {\color{black} and \textit{GLPdepth}}. This dataset includes some of the sequences where neural networks have been trained as well as others with similar characteristics not used for training, following the Eigen split proposal \cite{eigen2014depth}. This contains challenging sequences for pure monocular odometries since there exist almost pure turns which lead to big scale drift. In addition, it has a low frame rate (10 fps) which makes tracking more difficult, and has important luminosity changes which may cause brightness affine parameters to easily diverge. {\color{black} All experiments have been run in a high performance laptop. Inference time for \textit{monodepth2} on a RTX2080 GPU is 63ms, while tracking time per frame is on average 128ms. Since depth prediction is only needed for keyframes, this makes our approach work at $\sim$7 frames per second if not parallelization is used. Inference time for \textit{GLPdepth} is significantly higher, $\sim$144 ms.}

\setlength\tabcolsep{2pt}
\begin{table}
\centering
\caption{RMSE ATE (m) errors for KITTI odometry dataset.}
\begin{tabular}{c|c|c||c|c|c||c|}
\cline{2-7}
& \multicolumn{2}{c||}{Pure monocular} &  \multicolumn{3}{c||}{Monocular learning-based} & \multicolumn{1}{c|}{Stereo} \\
\cline{2-7}
& \multicolumn{1}{c|}{\doble{ORB-SLAM}{(No LC)$\ddag$}} & \multicolumn{1}{c||}{DSO*} &  \multicolumn{1}{c|}{\doble{DF-VO}{\cite{zhan2020visual}}} & \multicolumn{1}{c|}{{\color{black}\triple{Ours}{\em GLP}{\cite{kim2022global}}}} & \multicolumn{1}{c||}{\triple{Ours}{\em monodepth2}{\cite{zhan2020visual}}} & \multicolumn{1}{c|}{\doble{ORB-SLAM}{(No LC)$\ddag$}} \\
\hline
\multicolumn{1}{|c|}{00} & 77.19 & 113.18 & 11.34 & {\color{black}10.52} & \textbf{7.10} & 3.99 \\
\multicolumn{1}{|c|}{01} & 110.12 & - & 484.86 & {\color{black}\textbf{17.12}} & 25.38 & 1.38 \\
\multicolumn{1}{|c|}{02} & 34.32 & 116.81 & 21.16 & {\color{black}31.16} & \textbf{14.12} & 8.82\\
\multicolumn{1}{|c|}{03} & 0.90 & 1.39 & 2.04 & {\color{black}3.41} & \textbf{1.71} & 0.25\\
\multicolumn{1}{|c|}{04} & 0.72 & 0.42 & 0.86 & {\color{black}0.31} & \textbf{0.29} & 0.22\\
\multicolumn{1}{|c|}{05} & 36.29 & 47.46 & \textbf{3.63} & {\color{black}5.61} &  7.51 & 2.18\\
\multicolumn{1}{|c|}{06} & 52.61 & 55.62 & \textbf{2.53} & {\color{black}6.23} & 4.05 & 1.81\\
\multicolumn{1}{|c|}{07} & 17.04 & 16.72 & \textbf{1.72} & {\color{black}3.45} & 2.41 & 1.43\\
\multicolumn{1}{|c|}{08} & 56.42 & 111.08 & \textbf{5.66} & {\color{black}9.30} & 10.43 & 3.22\\
\multicolumn{1}{|c|}{09} & 55.74 & 52.23 & 10.88 & {\color{black}11.32} & \textbf{9.15} & 3.26 \\
\multicolumn{1}{|c|}{10} & 8.44 & 11.09 & 3.72 & {\color{black}4.36} & \textbf{3.40} & 0.88 \\ \hline \hline
\multicolumn{1}{|c|}{Avg.$\dagger$} & {33.97} & 52.6 & 6.35 & \textcolor{black}{8.57} &  \textbf{6,02} & 2.60   \\ 
\hline
\multicolumn{7}{l}{\scriptsize *: Results for DSO are extracted from \cite{zhan2020visual}} \\
\multicolumn{7}{l}{\scriptsize $\dagger$: KITTI01 is not used for average to ease comparison between methods} \\
\multicolumn{7}{l}{\scriptsize {$\ddag$: For ORB-SLAM we use the latest implementation from \cite{campos2021orb}} } \\

\end{tabular}
\label{tab:deepSLAM_comprative}
\end{table}

We compare our system against monocular DSO \cite{Engel-et-al-pami2018}, monocular and stereo versions of ORB-SLAM \cite{mur2015orb, mur2017orb} , with the loop-closing thread deactivated, for a more fair comparison, and also against DF-VO \cite{zhan2020visual} a monocular odometry based on depth prediciton. We measure the RMS of Absolute Trajectory Error (ATE) \cite{sturm2012benchmark} which are reported at table \ref{tab:deepSLAM_comprative}. Compared with pure monocular systems, our method based on depth prediction has an accuracy 5 times higher than ORB-SLAM and 9 times higher than DSO. Compared with the similar system DF-VO, which is the closest system to ours also using depth prediction, our proposal {\color{black} using \textit{monodepth2}} achieves a higher accuracy for 7 out of 11 sequences, with a lower average error. We use bold characters for better results among this comparison. Using a stereo systems gives better performance than our system, with a 57\% error reduction. {\color{black} Comparing our two solutions based on different depth-prediction networks, we see that \textit{monodepth2} based one systematically outperforms \textit{GLPdepth} in most of datastets, being the high-way environment sequence 02 the only exception. During experiments we detected that \textit{GLPdepth} was leading to more inconsistent scale prediction along consecutive frames than \textit{monodepth2}, despite of being in practice more precise}.

\begin{figure}
    \begin{subfigure}{.49\columnwidth}
      \centering
      \includegraphics[width=1.0\linewidth]{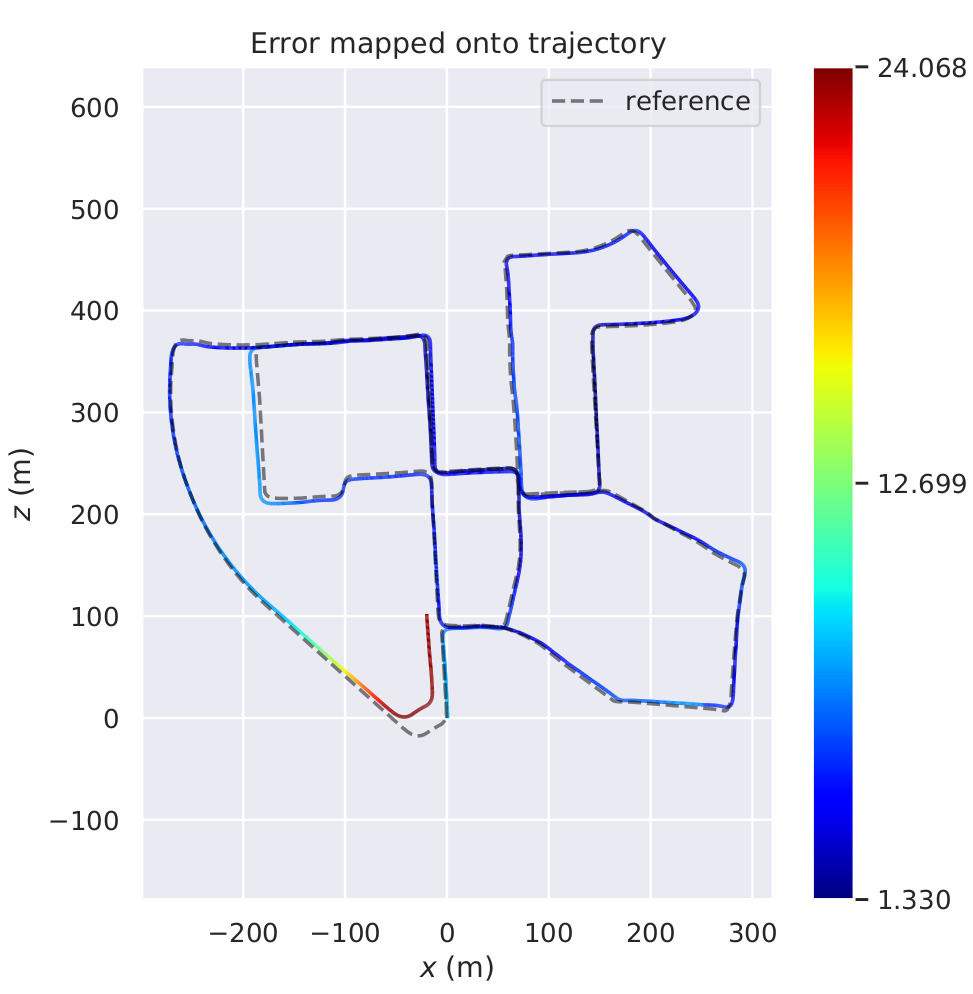}
      \caption{KITTI00}
      \label{subfig:KITTI0_map}
    \end{subfigure}
    \begin{subfigure}{0.49\columnwidth}
      \centering
      \includegraphics[width=1.0\linewidth]{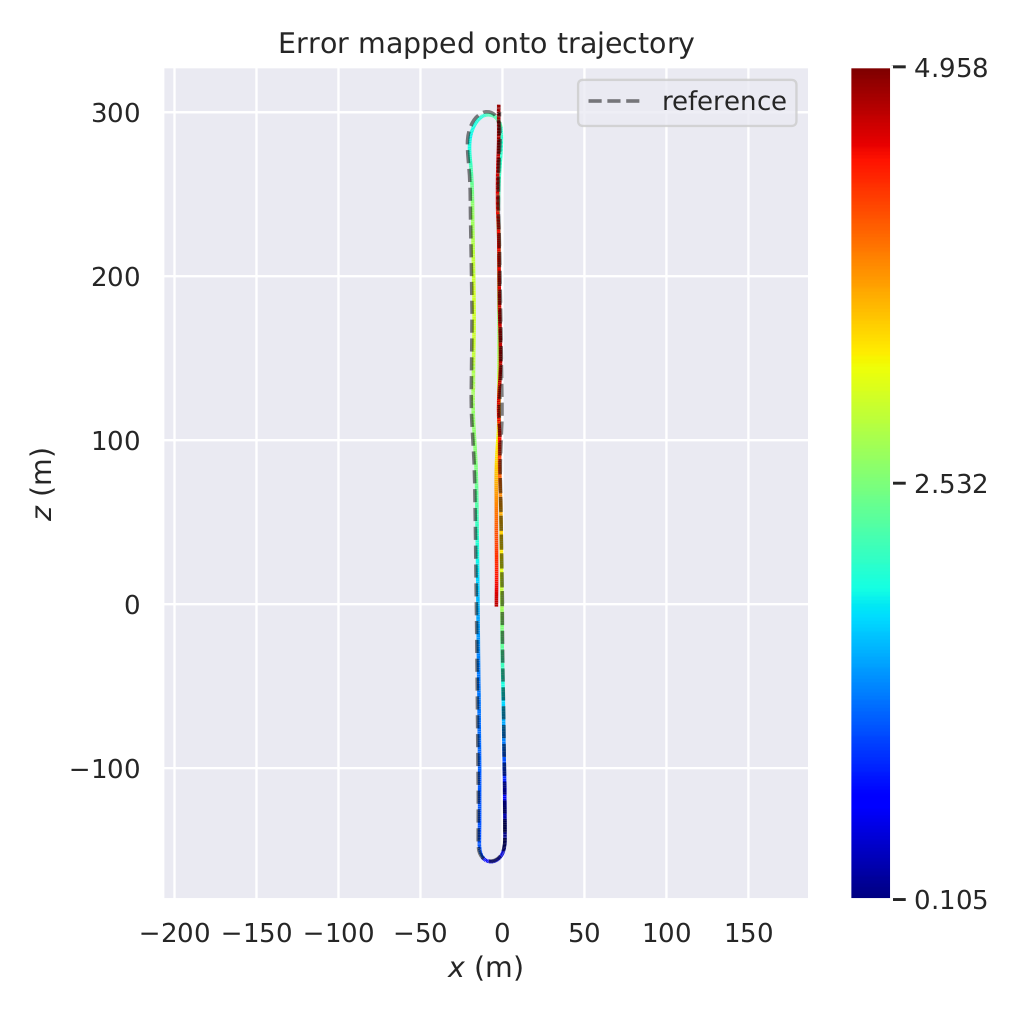}
      \caption{KITTI06}
      \label{subfig:KITTI6_map}
    \end{subfigure}
    
    \begin{subfigure}{0.49\columnwidth}
      \centering
      \includegraphics[width=1.0\linewidth]{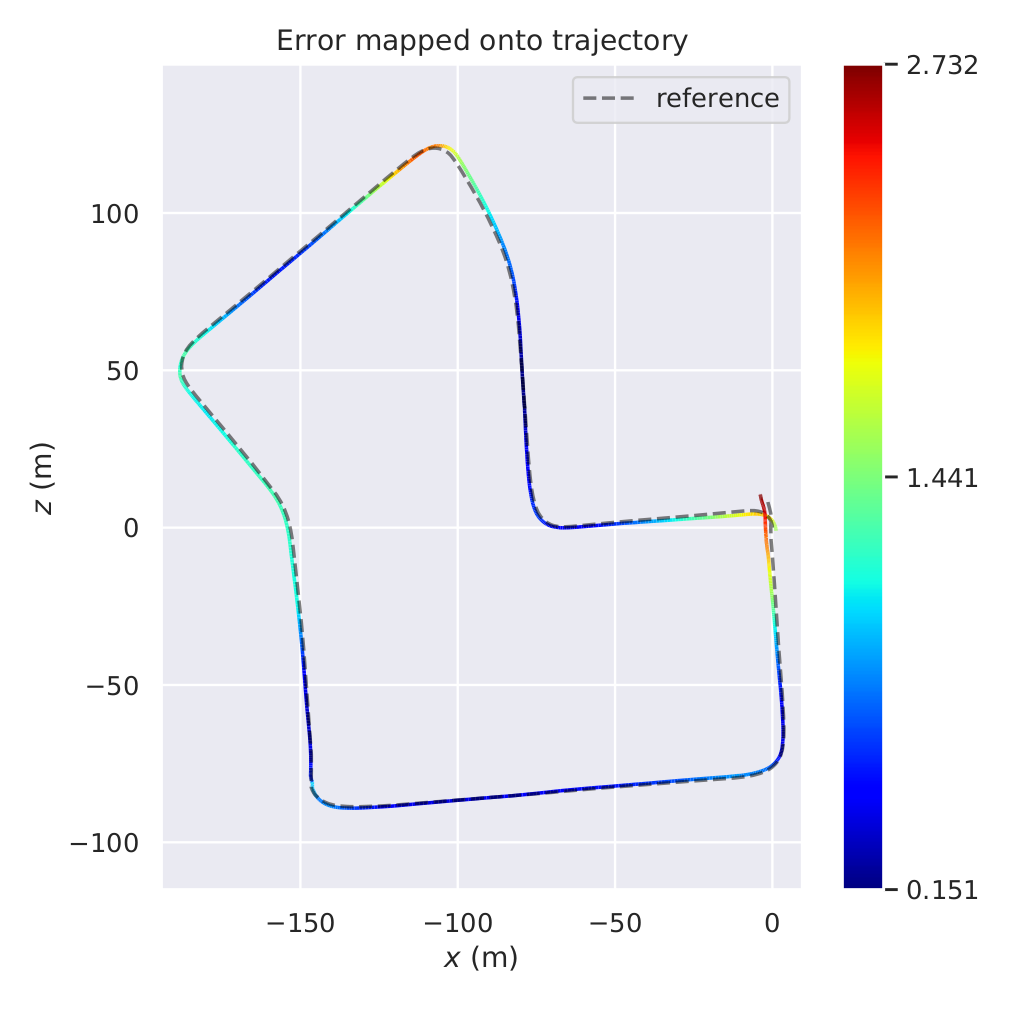}
      \caption{KITTI07}
      \label{subfig:KITTI7_map}
    \end{subfigure}
    \begin{subfigure}{0.49\columnwidth}
      \centering
      \includegraphics[width=1.0\linewidth]{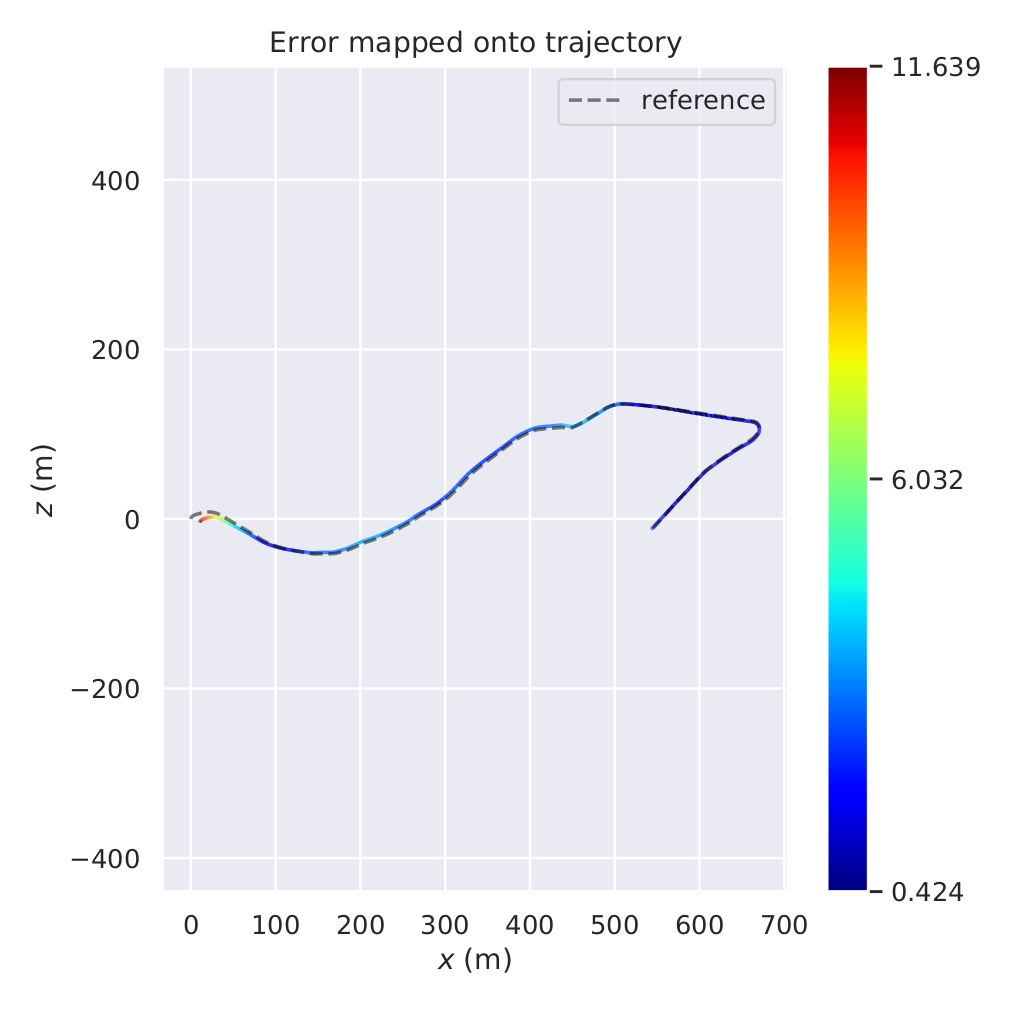}
      \caption{KITTI10}
      \label{subfig:KITTI10_map}
    \end{subfigure}
    \caption{Some sample trajectories {\color{black} using \textit{monodepth2}} at KITTI dataset coloured by ATE error with respect to ground-truth. 
    }
    \label{fig:KITTI_trajectories}
\end{figure}

\begin{figure}
\centering
    \begin{subfigure}{0.49\columnwidth}
      \centering
      \includegraphics[width=1.0\textwidth]{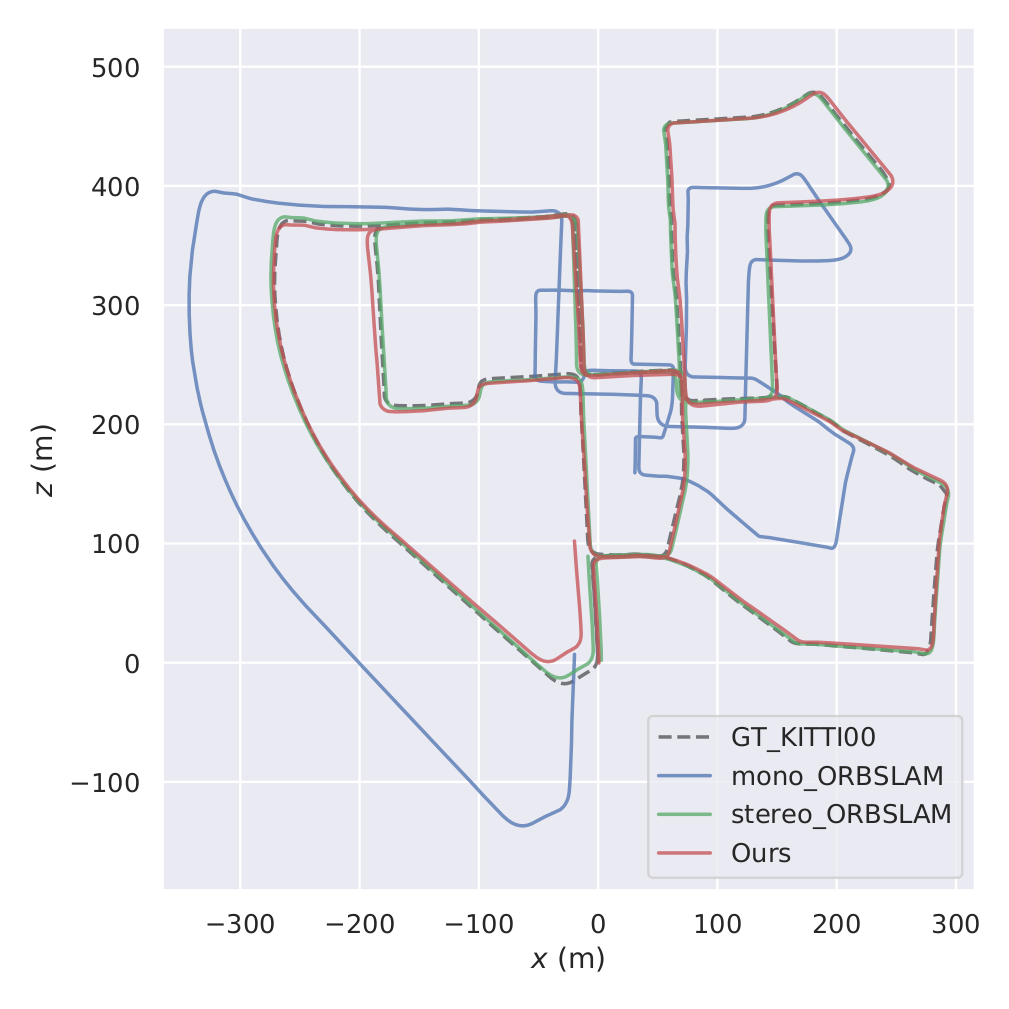}
      \caption{Aligned trajectories}
      \label{subfig:KITTI0_comparative}
    \end{subfigure}
    \begin{subfigure}{0.49\columnwidth}
      \centering
      \includegraphics[width=1.0\textwidth]{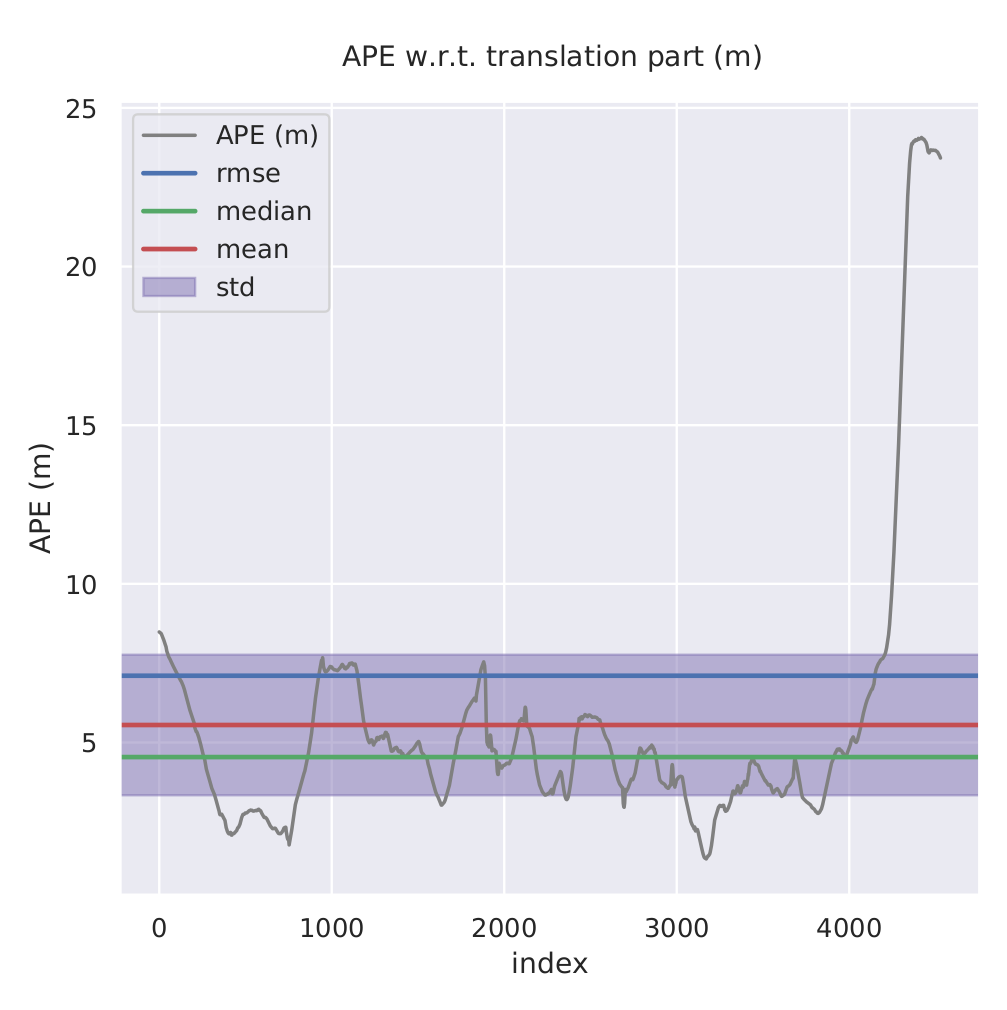}
      \caption{ATE evolution for our method}
      \label{subfig:KITTI0_ate_evolution}
    \end{subfigure}

    \caption{Comparative results for KITTI00 dataset. Notice how our proporsal does not suffer scale drift.}
    \label{fig:KITTI00_comparative}
\end{figure}

Several results of our proposal are shown in figure \ref{fig:KITTI_trajectories}. Major source of error for monocular odometries, scale drift, has completely disappeared leading to much more accurate results. More detailed results for sequence 00 are presented in figure \ref{fig:KITTI00_comparative}. 
We highlight that despite the lack of a loop closing module in our solution, we achieve zero drift for most of this sequence, as shown in figure \ref{subfig:KITTI0_comparative}, where multiple paths along the same streets can barely be differentiated. In this sense, regarding figure \ref{subfig:KITTI0_ate_evolution}, we can see that our proposal accumulates most of the error in a small section of the trajectory at the end of the sequence. In fact, this part corresponds with a non-urban environment, where both sides of the road are covered with vegetation, leading to less accurate results from \textit{monodepth2} as previously hypothesized. If this last part were not considered, the RMS ATE would be much closer to the median ATE, equal to 4.63m. An example of the reconstructed point cloud, as well as the sparse depth map used for frame tracking (see section \ref{sec:photometric_tracking}) are shown in figure \ref{fig:teaser_image}, top page.

For a more precise solution, a more complex neural network could be used as other methods do, but the solution would be less general. At DVSO paper \cite{yang2018deep}, they report results obtained with its odometry (DSO) using \textit{monodepth2} for the depth prediction, obtaining results with an accuracy close to ours. On the other hand, D3VO gets impressive results, but they do not report results using depth prediction from other neural networks and their implementation is not open-source and cannot be adopted in our work.

In addition, for the supervised training of these two works, they use a stereo version of DSO, which computes a very accurate true scale point cloud for high gradient points. Its neural networks learn to estimate very precisely this kind of points, which are also used along their odometries based on DSO. Coupling the odometry and the depth prediction in this way probably reduces the generality, which is one of our goals, but leads to very accurate results.

\section{Conclusions}
In this work we have presented a direct monocular odometry based on depth prediction from neural networks. We have shown that combining multi-view depth prediction and photometric residuals in a single optimization makes scale observable, removes scale-drift and leads to a much more accurate estimation than pure monocular solutions. Using a truncated robust cost (\textit{TLS}) for depth residuals allows us to consider only consistent measurements, making our optimization robust against spurious depth data. Our solution only requires the predicted depth estimation from the neural network, making our solution very general and enabling to integrate it along with most of existing networks. {\color{black} Taking two depth prediction neural networks as they are, \textit{monodepth2} and \textit{GLPdepth}, we have shown how our approach is suitable to work with different existing solutions, just plugging them into our pipeline}. We have also evaluated our system in multiple sequences and compared against monocular SLAM and DF-VO, system similar to ours. Our system gets results 5 to 9 times more precise than monocular solutions and is more accurate than DF-VO for 64\% of the sequences. The experiments demonstrate the validity of our proposal.

As future work we identify updating this odometry to a complete SLAM system, with a covisible and not temporal optimization window, which would boost its performance. This would entail some unsolved challenges for long term SLAM since images, depth prediction and its gradients and pyramidal decomposition should be stored in memory. In addition, including a loop-closing module would also be a big improvement. Making it work directly on images, without using features, is also an open problem.

\appendix
In this appendix we show how our proposed novel depth-prediction residual closely relates with photometric one, not supposing an important computational increment with respect the only photometric optimization. We provide derivatives for both errors, showing they have multiple common terms.

\vspace{3mm}
\textit{1. Photometric error}
\vspace{1.mm} 

For clarity sake we assume $\mathcal{N}_{\mathbf{u}_j} \equiv \mathbf{u}_j$. Derivatives for affine parameters $(a,b)$ are trivial, we omit them.
For $\rho^j$,
\begin{align}
    \frac{\partial r^j_{i,\text{photo}}}{\partial \rho^j} & = -\frac{e^{a_h}}{e^{a_i}} \frac{\partial I_i(\mathbf{u}_i^j)}{\partial \rho^j} \nonumber \\
    & = -\frac{e^{a_h}}{e^{a_i}} \left. \frac{\partial I_i}{\partial \mathbf{u}} \right|_{\mathbf{u}_i^j} \left. \frac{\partial \pi}{\partial \mathbf{x}}\right|_{\rho^j \mathbf{x}^j_i} \frac{\partial(\mathbf{R}_{ih}\mathbf{\bar{x}}^j_h  + \rho^j\mathbf{t}_{ih} )}{\partial \rho^j} \nonumber 
\end{align}
\begin{equation}
    \label{eq:photo_deriv_wrt_idepth}
    \boxed{\frac{\partial r^j_{i,\text{photo}}}{\partial \rho^j} = -\frac{e^{a_h}}{e^{a_i}} \underbrace{\left. \frac{\partial I_i}{\partial \mathbf{u}} \right|_{\mathbf{u}_i^j}}_{\text{Photometric}} \underbrace{\left. \frac{\partial \pi}{\partial \mathbf{x}} \right|_{\rho^j \mathbf{x}^j_i} \mathbf{t}_{ih}}_{\text{Geometric}}} \in \mathbb{R}
\end{equation}
where $\mathbf{\bar{x}}^j_h=\mathbf{x}^j_h/\lVert\mathbf{x}^j_h\lVert$ is fixed, $\partial I/\partial \mathbf{u}$ is the image gradient and $\partial \pi / \partial \mathbf{x}$ the camera projection derivative. For host pose, applying a small update $\xi_h \in \mathfrak{se}(3)$ in the world reference
\begin{equation}
\label{eq:photo_der_host_pose_prev}
    \frac{\partial r^j_{i,\text{photo}}}{\partial \xi_h} = -\frac{e^{a_h}}{e^{a_i}} \left. \frac{\partial I_i}{\partial \mathbf{u}} \right|_{\mathbf{u}_i^j} \left. \frac{\partial \pi}{\partial \mathbf{x}} \right|_{\rho^j \mathbf{x}^j_i} \left. \rho^j \frac{\partial ( \mathbf{x}_i^j)}{\partial \xi_h}\right|_{\mathbf{0}}
\end{equation}
where\\
\resizebox{\columnwidth}{!}{
  \begin{minipage}{\linewidth}
\begin{align}
    \left. \frac{\partial \mathbf{x}_i^j}{\partial \xi_h}\right|_{\mathbf{0}} & = \left. \frac{\partial \mathbf{T}_{i}  ( \mathbf{T}_{h}  \text{Exp}(\xi_h))^{-1}  \mathbf{x}_h^j }{\partial \xi_h} \right|_{\mathbf{0}} = \left. \frac{\partial (\mathbf{T}_{i}  \text{Exp}(-\xi_h)  \mathbf{T}_{h}^{-1}  \mathbf{x}_h^j )}{\partial \xi_h} \right|_{\mathbf{0}} \nonumber \\
     & = \left. \frac{\partial \text{Exp}(-\textbf{Ad}_{\mathbf{T}_{i}} \xi_h)  \mathbf{x}_i^j}{\partial \xi_h} \right|_{\mathbf{0}} = - (\mathbf{I}_3 | - [\mathbf{x}_i^j]_{\times})\textbf{Ad}_{\mathbf{T}_{i}}  \nonumber
\end{align}
  \end{minipage}
}

Where we used the adjoint definition \cite{sola2018micro}. This leads to
\begin{equation}
\label{eq:photo_der_host_pose}
    \boxed{\frac{\partial r^j_{i,\text{photo}}}{\partial \xi_h} = \rho^j \frac{e^{a_h}}{e^{a_i}} \left. \frac{\partial I_i}{\partial \mathbf{u}} \right|_{\mathbf{u}_i^j} \left. \frac{\partial \pi}{\partial \mathbf{x}} \right|_{\rho^j \mathbf{x}_i^j}  (\mathbf{I}_3 | - [\mathbf{x}_i^j]_{\times})\textbf{Ad}_{\mathbf{T}_{i}}}
\end{equation}
Same procedure for the observer frame $i$ leads to the same results but with different sign. This similarity between host and observer derivatives, which does not exist for local updates, is exploited when computing Hessian, yielding an important saving.

\vspace{3mm}
\textit{2. Depth-prediction error}
\vspace{1.mm}

Depth prediction residual $r_{i,\text{depth}}^j$ (\ref{eq:depth_pred_res}) shares common terms with photometric cost. For inverse depth, similar to (\ref{eq:photo_deriv_wrt_idepth}):
\begin{equation}
    \frac{\partial r_{i,\text{depth}}^j}{\partial \rho^j} = \left. \frac{\partial D^i_{NN}}{\partial \mathbf{u}}\right|_{\mathbf{u}^j_i} \left. \frac{\partial \pi}{\partial \mathbf{x}} \right|_{\rho^j \mathbf{x}^j_i} \mathbf{t}_{ih} - \frac{\partial \rho^j_i}{\partial \rho^j} 
\end{equation}
we now compute ${\partial \rho^j_i}/{\partial \rho^j}$:
\begin{align}
    \frac{\partial \rho^j_i}{\partial \rho^j} & = \frac{\partial [\mathbf{x}_i^j]_z^{-1}}{\partial \mathbf{x}_i^j} \frac{\partial \mathbf{x}_i^j}{\partial \rho_i^j} = (0,0,-[\mathbf{x}_i^j]_z^{-2}) \frac{\partial \left(\mathbf{R}_{ih} \mathbf{\bar{x}}^j_h/\rho^j + \mathbf{t}_{ih} \right)}{\partial \rho^j} \nonumber 
\end{align}
Finally leading to,
\begin{equation}
\label{eq:depth_pred_der_rho}
    \boxed{\frac{\partial r_{i,\text{depth}}^j}{\partial \rho^j} = \left. \frac{\partial D^i_{NN}}{\partial \mathbf{u}}\right|_{\mathbf{u}^j_i} \left. \frac{\partial \pi}{\partial \mathbf{x}} \right|_{\rho^j \mathbf{x}^j_i} \mathbf{t}_{ih} - \left(\frac{\rho_i^j}{\rho^j}\right)^2 [\mathbf{R}_{ih} \mathbf{\bar{x}}^j_h]_z}
\end{equation}
For host keyframe, as done for (\ref{eq:photo_der_host_pose}),

\resizebox{.9\linewidth}{!}{
  \begin{minipage}{\linewidth}
\begin{align}
    \frac{\partial r_{i,\text{depth}}^j}{\partial \xi_h} = \rho^j \left. \frac{\partial D^i_{NN}}{\partial \mathbf{u}}\right|_{\mathbf{u}^j_i} \left. \frac{\partial \pi}{\partial \mathbf{x}} \right|_{\rho^j \mathbf{x}^j_i} (\mathbf{I}_3 | - [\mathbf{x}_i^j]_{\times})\textbf{Ad}_{\mathbf{T}_{i}} - \frac{\partial \rho^j_i}{\partial \xi_h} 
\end{align}
  \end{minipage}}
  
Now, we compute ${\partial \rho^j_i}/{\partial \xi_h}$. We follow very similar procedure to (\ref{eq:photo_der_host_pose_prev}):

\begin{equation}
    \frac{\partial \rho^j_i}{\partial \xi_h} = \frac{\partial [\mathbf{x}_i^j]_z^{-1}}{\partial \mathbf{x}_i^j} \frac{\partial \mathbf{x}_i^j}{\partial \xi_h} = (0,0,[\mathbf{x}_i^j]_z^{-2}) (\mathbf{I}_3 | - [\mathbf{x}_i^j]_{\times})\textbf{Ad}_{\mathbf{T}_{i}} \nonumber
\end{equation}

This finally leads to:

\resizebox{.82\linewidth}{!}{
  \begin{minipage}{\linewidth}
\begin{equation}
\label{eq:depth_pred_der_host_pose}
    \boxed{\frac{\partial r_{i,\text{depth}}^j}{\partial \xi_h} = \left( \rho^j \left. \frac{\partial D^i_{NN}}{\partial \mathbf{u}}\right|_{\mathbf{u}^j_i} \left. \frac{\partial \pi}{\partial \mathbf{x}} \right|_{\rho^j \mathbf{x}^j_i} - (0,0,{\rho_i^j}^2) \right)  (\mathbf{I}_3 | - [\mathbf{x}_i^j]_{\times})\textbf{Ad}_{\mathbf{T}_{i}}}
\end{equation}
\end{minipage}}

For observer pose we can follow similar steps. From (\ref{eq:depth_pred_der_rho}) and (\ref{eq:depth_pred_der_host_pose}) we see that the only extra computation with respect to photometric residual comes from computing depth prediction image derivative $\partial D_{NN}^i / \partial \mathbf{u}$. All other terms are shared and do not yield a computational cost increment while Hessian sparse structure remains the same.

\addtolength{\textheight}{-11.9cm}   

\bibliographystyle{IEEEtran}
\bibliography{references,IEEEabrv}

\end{document}